\documentclass[sigconf]{acmart}

\usepackage{caption}
\usepackage[font=small,skip=0pt]{caption}
\usepackage{subcaption}
\usepackage{colortbl}
\usepackage{balance}
\usepackage{multirow}
\usepackage{pifont}
\usepackage{amsfonts}

\usepackage{bm}
\usepackage{algorithm}

\usepackage{enumitem}
\setlist[itemize]{leftmargin=*}

\AtBeginDocument{%
  }

\copyrightyear{2023}
\acmYear{2023}
\setcopyright{acmlicensed}\acmConference[CIKM '23]{Proceedings of the 32nd ACM International Conference on Information and Knowledge Management}{October 21--25, 2023}{Birmingham, United Kingdom}
\acmBooktitle{Proceedings of the 32nd ACM International Conference on Information and Knowledge Management (CIKM '23), October 21--25, 2023, Birmingham, United Kingdom}
\acmPrice{15.00}
\acmDOI{10.1145/3583780.3614915}
\acmISBN{979-8-4007-0124-5/23/10}

\usepackage{xspace}

\DeclareMathOperator*{\argmax}{arg\,max}

\begin{document}

\newcounter{cnt}
\setcounter{cnt}{181}
\newcommand{\db}[1]{\addtocounter{cnt}{1}\ding{\thecnt} \textbf{#1}}


\newcommand{\titlen}{Homophily-enhanced Structure Learning for Graph Clustering\xspace}
\newcommand{\titlename}{\textbf{ho}mophily-enhanced structure \textbf{le}arning for graph clustering\xspace}
\newcommand{\model}{HoLe\xspace}
\newcommand{\CSL}{Homophily-enhanced Structure Learning\xspace}

\title{\titlen}


\author{Ming Gu}
\orcid{0009-0005-2951-5256}
\affiliation{
  \institution{College of Computer Science and Technology, Zhejiang University}
  \country{China}
}
\email{guming444@zju.edu.cn}

\author{Gaoming Yang}
\orcid{0009-0008-2390-4092}
\affiliation{
  \institution{School of Software Technology, Zhejiang University}
  \country{China}
}
\email{ygm@zju.edu.cn}

\author{Sheng Zhou}
\orcid{0000-0003-3645-1041}
\affiliation{%
  \institution{Zhejiang Provincial Key Laboratory of Service Robot, Zhejiang University}
  \country{China}
}
\email{zhousheng_zju@zju.edu.cn}
\authornote{Corresponding author.}

\author{Ning Ma}
\orcid{0000-0001-7913-085X}
\affiliation{
  \institution{College of Computer Science and Technology, Zhejiang University}
  \country{China}
}
\email{ma_ning@zju.edu.cn}

\author{Jiawei Chen}
\orcid{0000-0001-7054-7974}
\affiliation{
  \institution{College of Computer Science and Technology, Zhejiang University}
  \country{China}
}
\email{sleepyhunt@zju.edu.cn}

\author{Qiaoyu Tan}
\orcid{0000-0001-8999-968X}
\affiliation{%
  \institution{Department of Computer Science, New York University Shanghai}
  \country{China}
}
\email{qiaoyu.tan@nyu.edu}

\author{Meihan Liu}
\orcid{0009-0006-1757-3368}
\affiliation{%
  \institution{College of Computer Science and Technology, Zhejiang University}
  \country{China}
}
\email{lmh_zju@zju.edu.cn}

\author{Jiajun Bu}
\orcid{0000-0002-1097-2044}
\affiliation{%
  \institution{College of Computer Science and Technology, Zhejiang University}
  \country{China}
}
\email{bjj@zju.edu.cn}


\renewcommand{\shortauthors}{Ming Gu et al.}


\begin{abstract}
Graph clustering is a fundamental task in graph analysis, and recent advances in utilizing graph neural networks (GNNs) have shown impressive results. Despite the success of existing GNN-based graph clustering methods, they often overlook the quality of graph structure, which is inherent in real-world graphs due to their sparse and multifarious nature, leading to subpar performance. Graph structure learning allows refining the input graph by adding missing links and removing spurious connections. However, previous endeavors in graph structure learning have predominantly centered around supervised settings, and cannot be directly applied to our specific clustering tasks due to the absence of ground-truth labels. To bridge the gap, we propose a novel method called \titlename~(\model). Our motivation stems from the observation that subtly enhancing the degree of homophily within the graph structure can significantly improve GNNs and clustering outcomes. To realize this objective, we develop two clustering-oriented structure learning modules, i.e., hierarchical correlation estimation and cluster-aware sparsification. The former module enables a more accurate estimation of pairwise node relationships by leveraging guidance from latent and clustering spaces, while the latter one generates a sparsified structure based on the similarity matrix and clustering assignments. Additionally, we devise a joint optimization approach alternating between training the homophily-enhanced structure learning and GNN-based clustering, thereby enforcing their reciprocal effects. Extensive experiments on seven benchmark datasets of various types and scales, across a range of clustering metrics, demonstrate the superiority of HoLe against state-of-the-art baselines.
\end{abstract}

\ccsdesc[500]{Mathematics of computing~Graph algorithms}
\ccsdesc[500]{Information systems~Clustering}

\keywords{Graph Clustering, Graphs Structure Learning, Graph Neural Networks}


\maketitle

\section{Introduction}

\begin{figure}
    \centering
    
    \begin{subfigure}[]{0.49\columnwidth}
         \includegraphics[width=\textwidth]{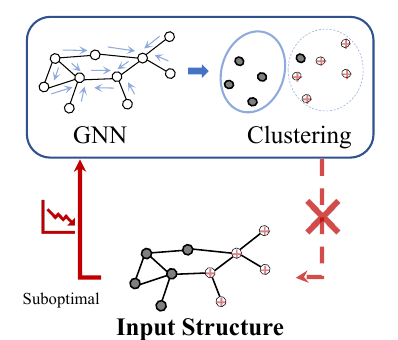}
         \caption{}
         \label{fig:motivation-v3-a}
    \end{subfigure}
    \begin{subfigure}[]{0.49\columnwidth}
         \includegraphics[width=\textwidth]{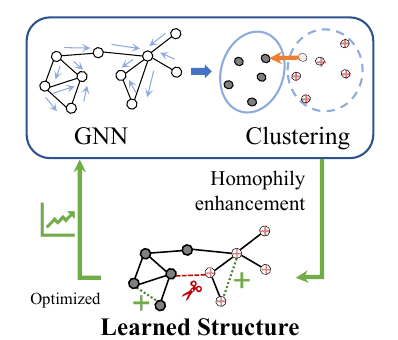}
         \caption{}
         \label{fig:motivation-v3-b}
    \end{subfigure}
    \caption{Comparison of GNN-based graph clustering methods. (a) Existing methods overlook the quality of the input structure, leading to suboptimal clustering outcomes. (b) Our approach integrates a clustering-oriented structure learning module, enhancing the structure quality and thereby improving clustering performance.}
    \label{fig:motivation}
    
\end{figure}

Graph data is ubiquitous in the real world and can express complex attributes and relationships between entities. 
Due to the inherent heterogeneity of this data, there is an underlying \textit{cluster structure} on the graph~\cite{schaeffer2007graph}, wherein nodes within the same cluster exhibit shared characteristics that differentiate from nodes in other clusters. 
This cluster structure holds great significance in understanding the organization of the graph, offering broad applications across diverse domains such as anomaly detection~\cite{ahmed2021graph}, molecular mining~\cite{grunig2022molecular}, face analysis~\cite{wang2019linkage}, and beyond. 
Unraveling the cluster structure of the nodes, commonly known as \textit{graph clustering}, has garnered substantial attention in recent years~\cite{yue2022survey} as a fundamental pursuit within graph analysis. Recently, graph neural network (GNN) based models have shown remarkable performance for graph clustering, capitalizing on GNN's ability to acquire informative representations by leveraging a graph's topological structure and rich attributes associated with its nodes.

Despite the notable progress achieved by existing GNN-based graph clustering methods, 
they often overlook the importance of the input graph's quality and only utilize the raw input graph structure throughout the entire clustering process.
\figureautorefname~\ref{fig:motivation-v3-a} illustrates the pipeline of existing methods.
However, suboptimal graph structures may arise due to the complex and contingent formation processes of graphs, which can be characterized by the potential absence of valuable links and the presence of spurious connections between nodes.
Take the social network as an example, a user may share similar preferences with substantial other users but can only connect with a limited subset due to unexposed relationships, leading to missing links. 
Moreover, two users may be connected simply because they are classmates, rather than indicating a shared preference, including spurious connections. The noisy nature of graphs, as acknowledged in social science~\cite{coscia2017network,wang2020learning}, not only affects social networks but also limits the expressive capacity of GNNs in node embedding, consequently compromising the performance of downstream tasks such as graph clustering. 

Recently, graph structure learning (GSL) has gained increasing prominence in the graph field~\cite{zhu2021deep,wang2021graph,liu2022towards}, owing to its capacity to acquire a more informative graph structure that supplants the original one, leading to improved results. 
Nonetheless, current research primarily emphasizes the utilization of labeled samples to facilitate structure learning~\cite{wang2021graph, chen2020iterative,fatemi2021slaps,franceschi2019learning,fu2022topology,jin2020graph}. 
These methods, unfortunately, do not readily extend to our specific clustering tasks due to the absence of ground-truth labels. 
Although the unsupervised graph structure learning~\cite{liu2022towards} has attracted growing interest recently and provides a potential solution by directly conducting GNN-based graph clustering on the learned structures, the lack of clustering-oriented optimization may result in suboptimal graph clustering results~\cite{bo2020structural, DFCN, SEComm}. \textit{Motivated by this research gap, in this paper, we aim to investigate how to employ graph structure learning to advance GNN-based graph clustering.} 

However, the challenges of integrating structure learning and GNN-based graph clustering are two-fold. \textbf{Firstly}, unlike structure learning with supervision~\cite{wang2021graph, chen2020iterative,fatemi2021slaps}, which relies on labeled samples to provide explicit signals for structure updates, graph clustering operates in an unsupervised manner without providing such dedicated guidance. 
Consequently, determining the optimal structure for the clustering task becomes arduous. 
\textbf{Secondly}, leveraging the clustering results to enhance structure learning presents another formidable challenge. Given the unsupervised nature of the task, it is unfeasible to treat all clustering outcomes as reliable guides. Thus, a fine-grained strategy is urgently required to distill the clustering results and facilitate effective structure refinement.

To tackle the aforementioned challenges, we propose a novel approach called \titlename (\model). 
Our empirical analysis of real-world datasets suggests that increasing the homophily of graph structures can benefit both GNNs and graph clustering. To achieve this, the \model method explicitly learns a homophily-enhanced structure during training, forming a mutually enhanced optimization loop, as depicted in \figureautorefname \ref{fig:motivation-v3-b}.
In particular, we aim to explore two critical research questions. (i) How to update the input structure and align the structure learning objective with the graph clustering tasks? (ii) How to effectively leverage the intermediate clustering results to advance structure updating during the training process, thereby promoting a mutually reinforcing effect between the two components. We summarize our main contributions as follows: 
\begin{itemize}
    \item We study how to utilize structure learning to improve the GNN-based graph clustering and provide the first homophily-enhanced proposal-HoLe for graph clustering. Specifically, we empirically demonstrate the efficacy of enhancing the degree of homophily within the graph structure in boosting clustering outcomes.   
    \item Motivated by our analysis, we propose hierarchical correlation estimation and cluster-aware sparsification modules, which redesign two critical components of the standard structure learning approach. These modules aim to improve the homophily of the learned structure and enhance its effectiveness. 
    \item We perform extensive experiments on 7 benchmark datasets to evaluate the superiority of our proposal. Notably, HoLe generally outperforms various state-of-the-art baselines across three evaluation metrics.
\end{itemize}

\section{Related Work}
\label{sec:relatedwork}

\textbf{GNN-based Graph Clustering}.
The remarkable expressive capabilities of GNNs have led to the popularity of GNN-based graph clustering in recent years~\cite{liu2022graph,yue2022survey}.
While the GNN backbone, such as graph convolutional networks~(GCNs)~\cite{kipf2016semi}, is not the primary focus of this work, we refer readers to~\cite{wu2020comprehensive,zhou2020graph} for a comprehensive view of advanced GNN models.
Regarding GNN-based graph clustering methods, they can be broadly categorized into two-stage and one-stage methods.
On the one hand, two-stage methods first learn node representations in an unsupervised manner and then perform clustering with various joint optimization strategies. For instance, many methods~\cite{DCRN,bo2020structural, DFCN, peng2021attention} exploit an auxiliary distribution as a guide following the self-training schema proposed by DEC~\cite{DEC}. 
On the other hand, one-stage methods simultaneously train representation learning and clustering with a joint loss for mutual improvement~\cite{SEComm, fettal2022efficient, gong2022attributed}. AGC-DRR~\cite{gong2022attributed} employs a clustering sub-network and proposes several constraints to form a one-stage contrastive training, while GCC~\cite{fettal2022efficient} minimizes the difference between the node representations and their reconstructed cluster representatives in a unified framework.
However, these methods overlook the significance of the graph topology, which may be suboptimal and limit the graph clustering performance.

\textbf{Graph Structure Learning}. 
Recently, there has been a surge in the development of methods addressing the graph structure learning (GSL) problem~\cite{zhu2021deep,zhou2023opengsl}.
The primary objective of GSL is to achieve a more refined and improved structure for graphs.
Typically, graph structure inference is performed to learn an adjacency matrix using available features in the absence of topology, while graph structure refinement is conducted to obtain a better adjacency relationship in the presence of noisy structures. 
The graph structure learning paradigms used can be divided into three categories. 
The first paradigm leverages certain metric learning functions, such as cosine similarity~\cite{chen2020iterative} and inner production~\cite{fatemi2021slaps}, to model structures as pairwise similarity measurements. 
The second paradigm exploits probabilistic models, e.g., Bernoulli probability model~\cite{franceschi2019learning} and stochastic block model~\cite{wang2021graph},  to formulate the structure learning procedure. 
The last one directly parameterizes the full graph into learnable parameters for optimization~\cite{fu2022topology, jin2020graph}. 
Although recent advancements in GSL, such as GEN~\cite{wang2021graph}, have witnessed a few works managing to model the cluster~(community) structure, it is important to note that these methods are primarily designed for supervised or semi-supervised settings and may not be directly applicable to unsupervised graph clustering scenarios.

\textbf{GNN-based graph clustering Meets Structure Learning}. 
Integrating unsupervised GSL with graph clustering seems to be an intuitive solution when it comes to enhancing the structure for graph clustering.
Although we have witnessed the growing interest in unsupervised GSL~\cite{liu2022towards}, the clustering information has not yet been incorporated into the structure learning process, which can not guarantee the enhancement of clustering on the learned structure.
Consequently, neither supervised nor unsupervised GSL methods are specifically tailored for the graph clustering task, and the learned structures from existing methods may not effectively enhance the clustering performance.

\section{Preliminaries}
\label{sec:preliminaries}

\subsection{Notations and Definitions}

Consider an undirected attribute graph $\mathcal{G}=\{\mathcal{V},\mathbf{X}, \mathcal{E}, \mathbf{A}\}$, where $\mathcal{V}=\{v_1, v_2, \cdots,v_N\}$ represents the node set, $\mathbf{X} \in \mathbb{R}^{N\times F}$ is the associated feature matrix, and $\mathcal{E}$ denotes the edge set. Here, $N$ represents the total number of nodes, while $F$ denotes the dimensionality of node features. The edges are represented by an adjacency matrix $\mathbf{A}=\{a_{ij}\}\in \mathbb{R}^{N\times N}$, where $a_{ij}=1$ if $(v_i, v_j)\in \mathcal{E}$, and 0 otherwise. Consequently, the degree matrix is denoted as $\mathbf{D}=\text{diag}(d_1,d_2,\cdots,d_N)\in \mathbb{R}^{N\times N}$, with $d_i=\sum^N_{j}a_{ij}$ representing the degree of node $v_{i}$. A commonly used definition of homophily, as introduced in~\cite{zhu2020beyond}, is presented in Definition \ref{def:homophily}.

\begin{definition}[Homophily]
\label{def:homophily}
Let $\mathcal{G}(\mathcal{V}, \mathcal{E}, \mathbf{X})$ represent a graph with $K$ disjoint classes, and denote the class assignment of node $v_i$ as $y_i$. Then the concept of edge homophily is characterized by the ratio of edges within the same class to the overall count of edges in the graph.
Formally, we have:
\begin{equation}
    \label{eqn:edge-homophily}
    h(\mathcal{G}, \{y_i;v_i \in \mathcal{V}\})=\frac{1}{|\mathcal{E}|}\sum_{(v_i, v_j)\in \mathcal{E}} \mathbb{I}(y_i=y_j),
\end{equation}
where $\mathbb{I}(\cdot)$ is an indicator function.
\end{definition}





The problem of graph clustering aims to group the node set $\mathcal{V}$ into $K$ disjoint clusters where each node $v_i$ is assigned to a certain cluster with nodes of similar patterns placed into the same one.
Denote $\textbf{Q}\in \mathbb{R}^{N\times K}$ as the cluster assignment matrix with $\textbf{Q}_{ik}$ meaning the probability of node $v_i$ belongs to the $k$-th cluster.
$\mathbf{C}=\argmax \mathbf{Q} \in \mathbb{R}^{N\times 1}$ represents the hard labels derived from the soft assignments $\textbf{Q}$.
$c_i$ is the $i$-th element of $\mathbf{C}$ denoting the hard cluster label of node $v_i$.

\subsection{GNN-based Graph Clustering}
\label{SGCL}
\textbf{GNN-based Node Representation Learning}. Most of the existing GNNs follow the message-passing schema that aggregates information from neighborhood nodes~\cite{zhou2020graph}.
We adopt a straightforward combination of neighborhood aggregation with linear transformation for our GNN backbone following the prior works~\cite{AGE,wu2019simplifying,dong2021equivalence},
which can be formulated as:
\begin{equation}
    \label{eqn:sgc}
    \mathbf{Z}=(\mathbf{I}-\kappa \mathbf{L})^{l}\mathbf{X}\mathbf{W},
\end{equation}
where $\mathbf{W}$ is the weight matrix of linear transformation, $l$ is the number of propagation times, $\kappa$ controls whether the filter is low-pass~\cite{AGE}, and $\mathbf{Z}$ is the output embedding. Given the graph structure $\mathbf{A}$,
$\mathbf{L}$ can be defined as $\mathbf{L}=\mathbf{I} - \mathbf{D}^{-\frac{1}{2}}   (\mathbf{\mathbf{A}} + \mathbf{I})  \mathbf{D}^{-\frac{1}{2}}$, where $\mathbf{I}$ is the identity matrix, and $\mathbf{D}$ is the degree matrix of $\mathbf{A}$. 

\begin{figure*}
    \centering
    \includegraphics[width=.85\textwidth]{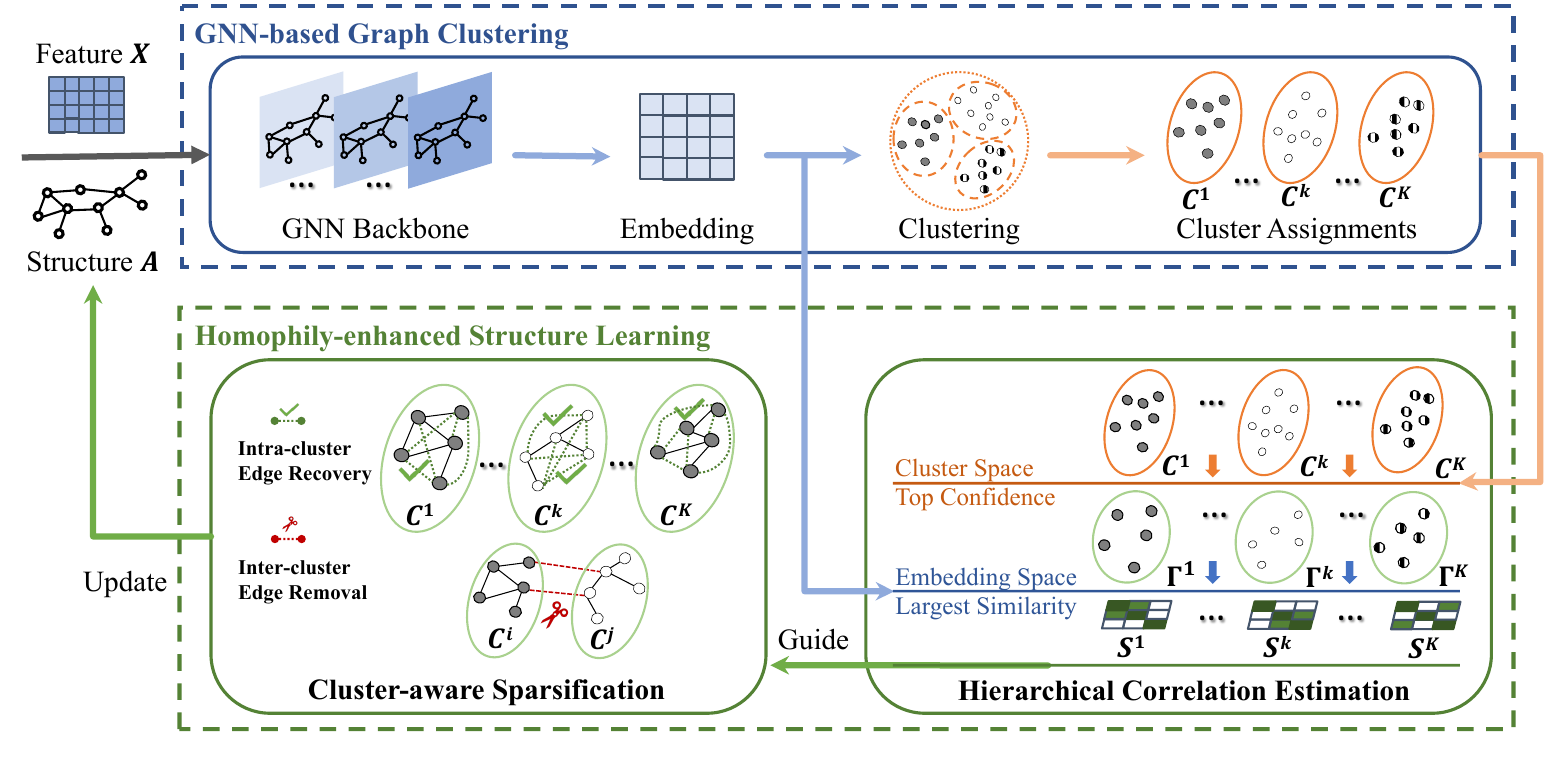}
    \caption{Overview of \model method. The homophily-enhanced structure learning module consists of hierarchical  correlation estimation and cluster-aware sparsification module, which takes explicit guidance from the embeddings and clustering results. This ensures the homophily enhancement of the learned structure, facilitating mutual reinforcement optimization with the GNN-based graph clustering.}
    \label{fig:framework}
\end{figure*}

\textbf{Graph Clustering}. The graph clustering module performs clustering on the node representations obtained from the GNN backbone. 
A widely used self-training clustering scheme~\cite{DEC,DCRN,DFCN,DAEGC} is to leverage the high-confidence results to improve subsequent training rounds through an auxiliary distribution. 
Specifically, the similarity between the node embedding $\mathbf{z_i}$ and the $k$-th cluster center $\bm{\mu_k}$ is measured by a Student's t-distribution: 
\begin{equation}
    \label{eqn:Q}
    q_{ik} = \frac{(1+\|\mathbf{z}_i - \bm{\mu}_k\|^2)^{-1}}{\sum_k^K(1+\|\mathbf{z}_i-\bm{\mu}_k\|^2)^{-1}},
\end{equation}
where $q_{ik}$ can be interpreted as the probability of assigning node $v_i$ to the $k$-th cluster (i.e., a soft assignment). 
Then the clustering distribution is optimized with the help of an auxiliary target distribution in a self-training way by minimizing the KL divergence~\cite{kullback1951information}:
\begin{equation}
    \label{eqn:cls}
    \mathcal{L}_{cls} = KL(P\|Q)=\sum_i \sum_k p_{ik}\log \frac{p_{ik}}{q_{ik}},
\end{equation}
where $p_{ik} = \frac{q_{ik}^2/\sum_j q_{jk}}{\sum_k^K(q_{ik}^2/\sum_j q_{jk})}$. 
As the auxiliary distribution $P$ raises the clustering assignment distribution $Q$ to the second power, this encourages $Q$ to be a sharp clustering assignment distribution with the guidance of high confidence points and avoid collapsing into a single cluster~\cite{DEC,zhou2022comprehensive}.

Please note that we only study the naive GNN backbone and clustering module that have been widely used in existing works, while other selections can also be adopted here and we leave them in our future works.

\section{Methodology}
In this section, we present an overview of the proposed HoLe, as depicted in Figure~\ref{fig:framework}. In Section~\ref{sec:data-analysis}, we empirically analyze our motivation to prioritize the homophily metric for structure learning in graph clustering scenarios. Then, we describe two tailored structure learning modules designed to enhance the degree of homophily in learned structure in Section~\ref{sec:gsl} and illustrate how to jointly optimize HoLe in Section~\ref{sec:joint}. Finally, in Section~\ref{sec:gsl-analysis}, we perform a sanity check to demonstrate the effectiveness of our proposal in enhancing the homophily of the learned structure.  

\subsection{Motivation}
\label{sec:data-analysis}

\label{sec:methodology}

As discussed in the introduction, designing effective structure learning techniques for advancing GNN-based graph clustering is challenging due to the absence of ground-truth labels.
One intuitive solution is to train the unsupervised structure learning module by approximating the original graph structure, as demonstrated in previous works~\cite{liu2022towards}. 
However, we argue that the learned structure may not be particularly beneficial for graph clustering. 
The primary reason is that the objective focuses on preserving the input graph in the hidden space, deviating from the requirements of graph clustering where nodes share similar clustering assignments that should be close in the embedding space.
To address this limitation, we suggest a novel approach by deducing the cause from the effect and identifying the type of structure that satisfies the clustering needs. 
Given that GNNs typically learn similar embeddings for nodes that are close in the graph structure~\cite{kipf2016semi,zhou2020dge,wu2019simplifying,zhang2018anrl,velickovic2017graph,peng2020graph}, an ideal structure for graph clustering might involve connecting nodes within the same cluster while removing connections between different clusters, with a particular emphasis on achieving a high degree of homophily. 
This approach would enable GNNs to learn highly discriminative embeddings, which are more conducive to clustering.

\textbf{Experiments Design}. To verify our hypothesis, we conduct experiments by manually controlling the homophily of the graph from an oracle perspective and assessing its impact on the clustering results. 
Specifically, we use the graph autoencoder (GAE)~\cite{kipf2016semi} as the embedding backbone and employ K-means~\cite{lloyd1982least} for the graph clustering task. 
Based on the definition of homophily, we develop two strategies, namely intra-class edge recovery (ERC) and inter-class edge removal (ERM), to increase the homophily degree of the input graph.
In the ERC strategy, we randomly select a proportion, denoted as $\gamma$, of nodes from each class and add new edges between pairs of nodes with the highest inner product. The number of added edges is determined by a percentage, denoted as $\xi$, relative to the original number of edges in the graph. Similarly, in the ERM strategy, we randomly remove a proportion, denoted as $\eta$, of edges that connect nodes from different classes. The values of $\gamma$, $\xi$, and $\eta$ are calculated based on the desired increase in homophily.

\begin{figure}[t]
    \centering
    \includegraphics[width=.4\textwidth]{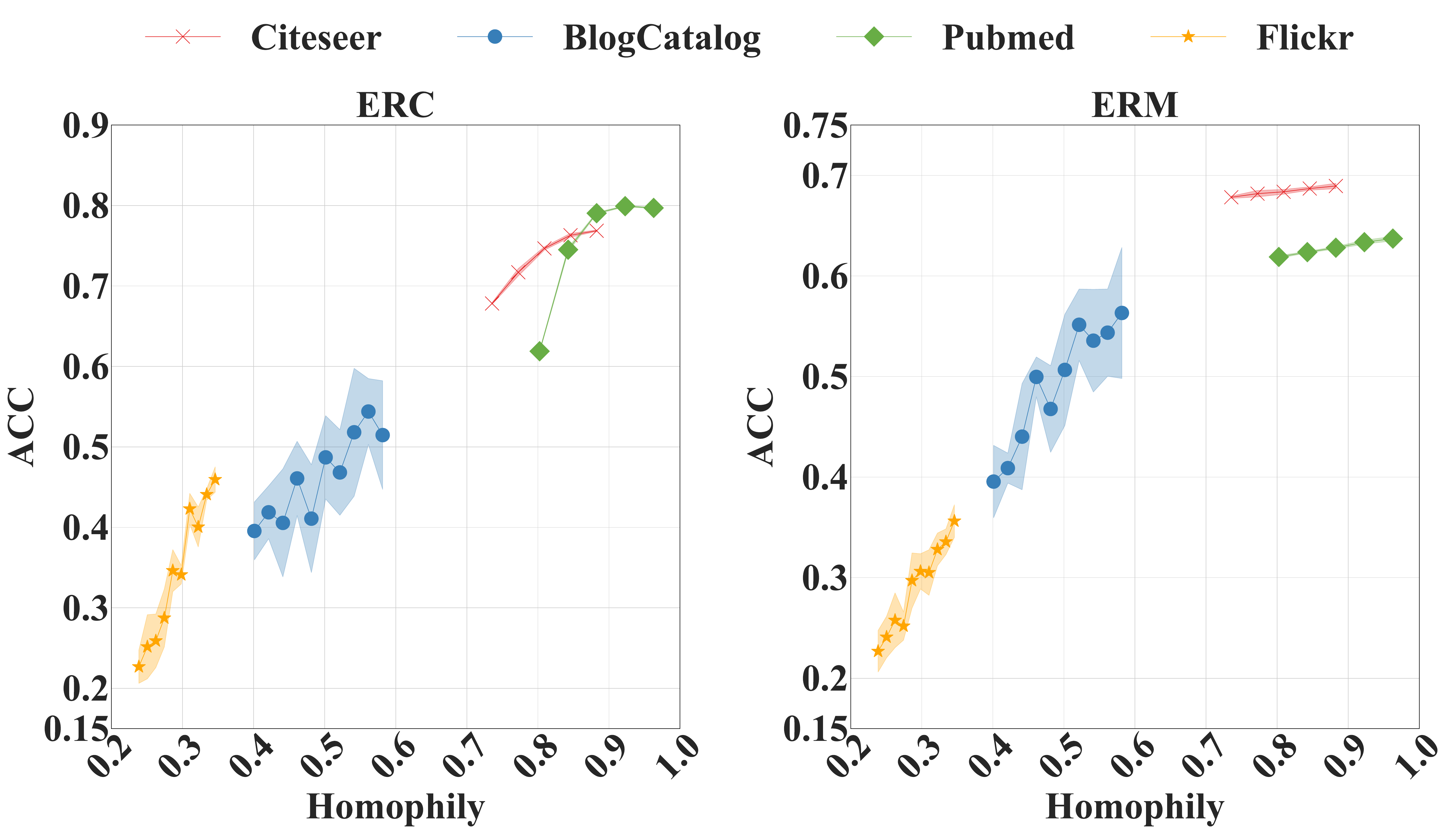}
    \caption{Empirical Analysis.
    The homophily of each dataset's points experiences an increment at regular intervals of 5\%.
    }
    \label{fig:data-analysis}
\end{figure}

\textbf{Observations}. 
Figure \ref{fig:data-analysis} illustrates the graph clustering results with respect to the homophily ratio, revealing two key observations. 
(1) \textbf{Increasing the homophily of structures using ERC and ERM consistently enhance the clustering performance across all four evaluated datasets}. The improvement demonstrates sustained and steady growth.
(2) \textbf{Even minor modifications in the homophily have significant impacts on the clustering outcomes}. 
The original homophily scores of the BlogCatalog and Flickr datasets, as shown in Table~\ref{tab:datasets}, are considerably lower compared to the other two datasets. Consequently, the structural updates in our experiment for these datasets are relatively smaller. However, their clustering improvements are significant. 

Based on the aforementioned observations, we argue that \textbf{the homophily metric serves as an effective indicator for improving graph clustering results}. 
However, enhancing the homophily ratio of the input graph in the graph clustering task is a non-trivial task, as we often lack ground-truth labels in practice.

\subsection{\CSL}
\label{sec:gsl}

Motivated by the observations presented in Section~\ref{sec:data-analysis}, we propose a principled approach, called HoLe, as depicted in Figure~\ref{fig:framework}, to enhance the homophily degree of the original graph structure in the context of graph clustering. 
The HoLe framework consists of two main learning branches: \textit{GNN-based graph clustering} and \textit{Homophily-enhanced structure learning}.
In the GNN-based graph clustering branch, we utilize a graph as input and obtain the clustering assignment matrix $\mathbf{Q}$ and intermediate node representation matrix $\mathbf{Z}$. 
The detailed design has been proposed in Section~\ref{SGCL}.

The main \textbf{innovations} of our approach lie in the lower branch of Figure~\ref{fig:framework}, where we aim to dynamically refine the homophily degree of the input structure during the training process. Unlike traditional structure learning methods that rely on labeled nodes, our homophily-enhanced structure learning approach utilizes the clustering results to facilitate structure refinement in an unsupervised learning setting. However, we encounter two major challenges: \textbf{C1)} obtaining a reliable estimation of node similarity in the clustering space. This difficulty arises primarily due to the presence of noise in the clustering results, especially during the early stages of training, owing to the unsupervised nature of the problem. \textbf{C2)} Even with a reliable node similarity matrix, identifying valuable missing links or eliminating spurious connections remains challenging due to the vast potential candidate space, which can be on the order of millions or even billions.
To tackle the above challenges, we carefully design the hierarchical correlation estimation and cluster-aware sparsification in the following subsections.

\subsubsection{Hierarchical Correlation Estimation} We start by discussing how to tackle the first \textbf{challenge C1}. Given the intermediate clustering results, the intuitive solution is to measure the pairwise correlation between nodes based on the clustering space, such as computing the inner product between two cluster vectors. However, this approach is suboptimal due to the presence of considerable noise in the clustering outcomes, particularly during the initial stages of training. To overcome this limitation, we propose a hierarchical correlation estimation module. The key concept is to enhance the estimation of node similarity by incorporating guidance from both the clustering and latent space in a hierarchical manner. Specifically, using the clustering assignments generated by the clustering algorithm, we first calculate the hard pseudo-label
: 
\begin{equation}
    c_{i}=\argmax_{k} q_{ik},
    \label{pseudo}
\end{equation}
where $q_{ik}$ denotes the probability of node $i$ being classified to the $k$-th cluster. After obtaining the pseudo-labels for all nodes, we select the top $\gamma$ percentage of nodes with the highest cluster assignment probability in each cluster $k$, resulting in a subset of nodes called the top-confident node subset $\Gamma^k$:
\begin{equation}
    \label{eqn:gmk}
    \Gamma^{k} = \{v_i|c_i=k, \text{Rank}|_{\{v_i|c_i=k\}}(q_{ik})\le \gamma * |\{v_i|c_i=k\}|\},
\end{equation}
where $\text{Rank}|_{\{v_i|c_i=k\}}(q_{ik})$ represents the ranking of cluster assignment probability of node $v_i$ within cluster $k$. By utilizing this approach, we can mitigate the impact of noisy clustering assignments to some extent. However, if we still rely on the cluster space to determine the edge weight, the presence of confident yet false assignments can still undermine the accuracy of the similarity matrix. To enhance the pairwise correlation between nodes, we propose estimating node similarity based on the latent space rather than the clustering space. 
Given the node embedding matrix $\mathbf{Z}$, the node similarity between all pairs of nodes can be measured by dot product $\mathbf{S} = \mathbf{Z} {\mathbf{Z}^T}$.
In particular, the node similarity between node pairs within the top-confident node subset $\Gamma^{k}$ can be expressed with their corresponding node embedding matrix $\mathbf{Z}^{k}$:

\begin{equation}
    \label{eqn:Sk}
    \mathbf{S}^{k}_{ij} = \mathbf{Z}^{k}_{i} {\mathbf{Z}^{k}_{j}}^T\in \mathbb{R}^{|\Gamma^k|\times |\Gamma^k|},
\end{equation}
where $|\Gamma^k|$ is the number of nodes in the subset $\Gamma^k$.

\underline{\textbf{Benefits}.} Diverging from conventional GSL approaches that solely rely on either soft assignments or the embedding similarity matrix to identify the missing or incorrect edges, our proposed method integrates both factors simultaneously. By incorporating the guidance of $\mathbf{S}^{k}$, we update the graph structure by considering the \textit{highest values of embedding similarity} within the subset of \textit{top-confident nodes in each cluster}. This approach significantly enhances the accuracy of the edge modification process, resulting in more precise and reliable graph structure learning.

\subsubsection{Cluster-aware Sparsification} After generating a more reliable node similarity matrix, we then discuss how to enhance the homophily degree of the learned structure (\textbf{challenge C2}). 
Instead of directly generating a sparsified structure by retaining the top-ranked node pairs based on the similarity matrix as existing works~\cite{liu2022towards,chen2020iterative,fatemi2021slaps}, we propose leveraging the underlying clustering patterns to guide the sparsification process with the aim of increasing the homophily degree. 
To be specific, we develop two cluster-aware strategies, namely \textit{intra-cluster edge recovery} and \textit{inter-cluster edge removal}, to respectively add intra-cluster edges and remove inter-cluster edges.

\textit{Intra-cluster Edge Recovery}. This strategy aims to add edges between nodes within a cluster. To fulfill the homophily requirement while maintaining the quality of added edges, we suggest selecting the top $\xi$ percent of pairs of nodes within each cluster $k$ that have the largest similarity $\mathbf{S}_{ij}^k$,  expressed as:
\begin{equation}
    \label{eqn:Ekrc}
    \mathcal{E}^{k}_{rc} = \{(v_i, v_j)\ |\ \text{Rank}(\mathbf{S}^{k}_{ij}) \leq \xi * |\mathcal{E}| * \frac{N_k}{N},\ \text{and}\ c_{i}=c_{j}=k\},
\end{equation}
where $N_{k}$ denotes the number of nodes in $\Gamma^{k}$, $\mathcal{E}$ is the existing edge set of the current structure learning round, and $\text{Rank}(\mathbf{S}^{k}_{ij})$ signifies the similarity ranking of node pair $(v_{i},v_{j})$ in descending order within $\mathbf{S}^{k}$. 
It is worth noting that self-loop edges or existing edges are not considered in the ranking process.
$\frac{N_k}{N}$ is used to ensure that the number of the selected edges is proportional to the number of nodes in the cluster $k$.
The complete set of recovered edges is obtained by merging sets from all clusters:
\begin{equation}
    \label{eqn:Erc}
    \mathcal{E}_{rc} = \bigcup_k^K \mathcal{E}^{k}_{rc},
\end{equation}
where $\bigcup$ indicates the union operation, and $\mathcal{E}_{rc}$ represents the final edge set after the union of $K$ clusters.

\textit{Inter-cluster Edge Removal}. In addition to adding new connections, this strategy aims to remove spurious edges that violate the homophily assumption from the existing edge set. For this purpose,
we select the bottom $\eta$ percentage of all existing edges in $\mathcal{E}$ that have the least similarity $\mathbf{S}_{ij}$ between clusters:

\begin{equation}
    \label{eqn:Erm}
    \mathcal{E}_{rm} = \left\{(v_i, v_j)\ |\ \text{Rank}(\mathbf{S}_{ij}) \geq (1-\eta) * |\mathcal{E}|,\ (v_i, v_j)\in\mathcal{E},\ c_i \neq c_j\right\},
\end{equation}
where $\mathcal{E}$ represents the edge set based on the current structure learning round. It is important to note that this step operates on all the existing edges in the entire graph, rather than focusing solely on a single cluster in the recovery operation.

\textbf{Structure Learning}. Given the recovered and removed edge sets (i.e., $\mathcal{E}_{rm}$ and $\mathcal{E}_{rc}$), the edge set associated with the current structure can be computed as $\mathcal{\widebar{E}}=\mathcal{E}-\mathcal{E}_{rm} + \mathcal{E}_{rc}$ in each training round. 
Formally, if we represent the corresponding adjacency matrices of $\mathcal{E}$, $\mathcal{E}_{rm}$ and $\mathcal{E}_{rc}$ as  $\mathbf{A}$, $\mathbf{A}_{\mathcal{E}_{rm}}$ and $\mathbf{A}_{\mathcal{E}_{rc}}$, the adjacency matrix $\mathbf{\widebar{A}}$ of the updated structure can be formulated as:

\begin{equation}
    \label{eqn:E'}
    \mathbf{\widebar{A}}=\mathbf{A}-\mathbf{A}_{\mathcal{E}_{rm}} + \mathbf{A}_{\mathcal{E}_{rc}}.
\end{equation}

In practice, the structure update process is usually conducted for multiple rounds, although our notations do not distinguish this for simplicity.
To ensure that the embedding captures the new proximity in the learned graph structure, we reconstruct the updated adjacency matrix in the latent space, expressed as:
\begin{equation}
    \label{eqn:reg}
    \mathcal{L}_{gsl}= \|\mathbf{Z}\mathbf{Z}^T-\mathbf{\widebar{A}}\|^2_F.
\end{equation}

Ideally, the homophily of the graph structure should increase during the training process.
This implies that in the learned graph structure $\mathbf{\widebar{A}}$, connected nodes are more likely to have similar labels compared to the original graph.
Consequently, by bringing the embeddings of connected nodes closer while pushing away those that are disconnected, we enhance the discriminability of the embeddings and thus improve the clustering performance.


\subsection{Joint Optimization}
\label{sec:joint}
Given the GNN-based graph clustering and homophily-enhanced structure learning, the overall objective of \model is formulated as:
\begin{equation}
    \mathcal{L}_{jnt} =  \mathcal{L}_{cls} + \mathcal{L}_{gsl}. 
\end{equation}
Here, $\mathcal{L}_{cls}$ and $\mathcal{L}_{gsl}$ represent the objectives of GNN-based clustering and homophily-enhanced structure learning, respectively.
To ensure the stability and mutual enhancement of training between these two components, we devise an iterative optimization loop that continues until the model converges or reaches the maximum structure learning epoch. 
During the inference stage, the label $c_i$ for node $v_i$ is determined by selecting the cluster with the largest probability, as given by Equation~\eqref{pseudo}.

Previous works~\cite{yue2022survey} have demonstrated the significant impact of parameter initialization, especially the clustering centers, on the performance of clustering algorithms. 
Therefore, in \model, before entering the optimization loop, we conduct training of the GNNs using the input graph structure, as described in Equations~\eqref{eqn:sgc} and \ref{eqn:reg}, to derive the node embeddings $\mathbf{Z}$. Subsequently, we initialize the cluster centers $\bm{\mu}_k$ by performing K-means on $\mathbf{Z}$. 
Afterward, the initial clustering assignments $\mathbf{Q}$ are obtained through the application of GNN-based graph clustering, as illustrated in Equation~\eqref{eqn:cls}.

\subsection{Sanity Check}
\label{sec:gsl-analysis}
In this section, we analyze the homophily enhancement achieved by our \model against the state-of-the-art unsupervised GSL method-SUBLIME. For all methods, we update the graph structure according to their own configurations. Then, we compute the homophily score of the learned structure as defined in Equation~\eqref{eqn:edge-homophily}.

\begin{figure}[t]
    \centering
    \includegraphics[width=.4\textwidth]{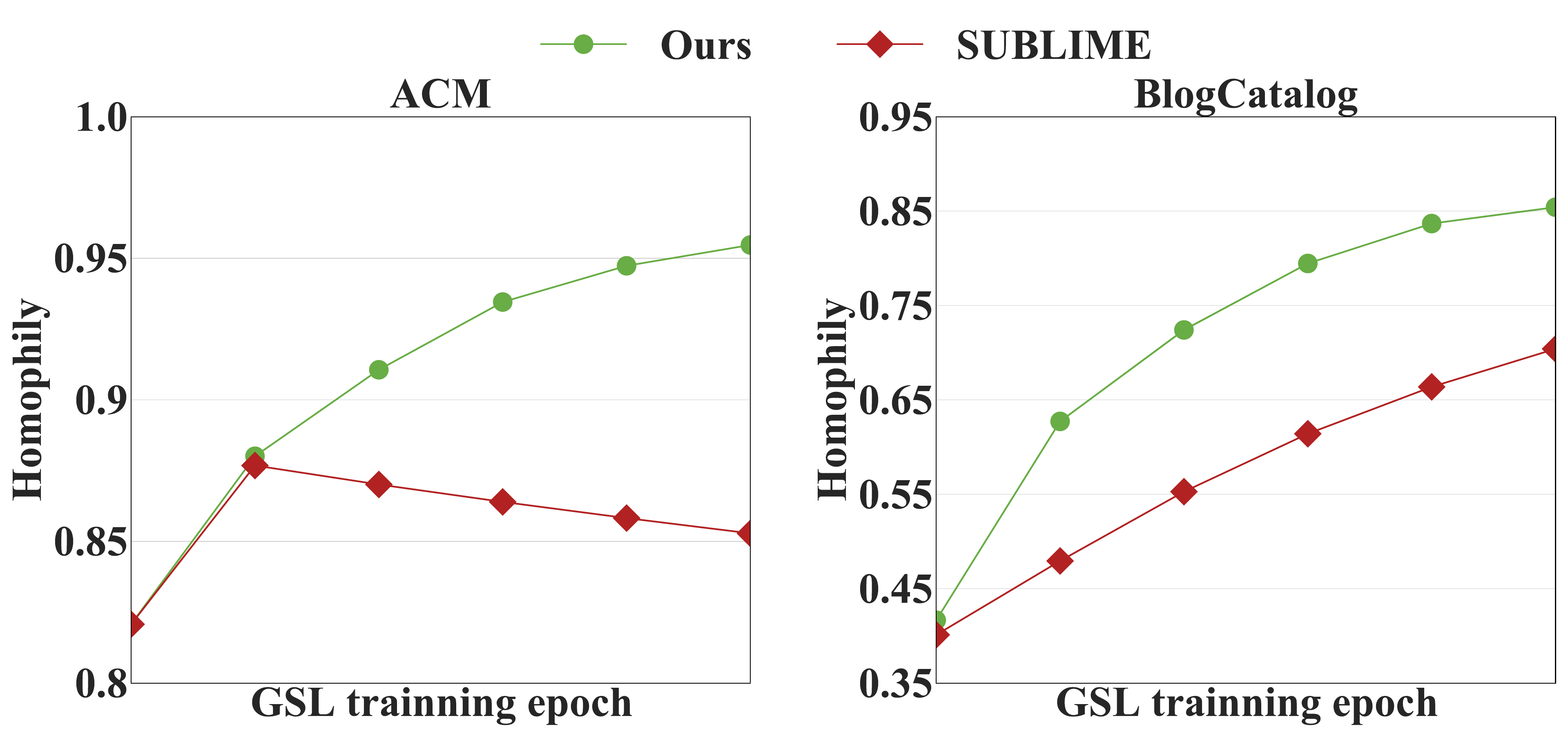}
    \caption{Quality of learned structures measured by homophily compared to unsupervised GSL method SUBLIME\cite{liu2022towards}.}
    \label{fig:structure-quality}
\end{figure}

Figure \ref{fig:structure-quality} reports the homophily results of HoLe and SUBLIME across the training epochs. It can be observed that the proposed HoLe exhibits a larger increase in homophily compared to SUBLIME. This is mainly because SUBLIME only focuses on approximating the original graph structure and does not leverage clustering outcomes to guide the structure learning process. Furthermore, as shown in Table \ref{tab:results}, HoLe outperforms SUBLIME across seven benchmark datasets on average. These results shed light on the effectiveness of HoLe in enhancing the homophily of the learned structure, confirming our motivation to explore homophily-enhance structure learning for graph clustering.

\section{Experiments}

In this section, we aim to answer the following research questions through extensive experiments on multiple real-world datasets:
\begin{itemize}
    \item \textbf{RQ1:} How does the proposed \model method perform compared to state-of-the-art graph clustering methods?
    \item \textbf{RQ2:} What are the contributions of different components in HoLe? 
    \item \textbf{RQ3:} What are the impacts of hyper-parameters on \model?
\end{itemize}

\begin{table*}[t]
    \centering
    \small
    \caption{Overall performance of graph clustering on seven datasets measured by ARI, NMI, and ACC scores in percentage. The best results are in bold, and the second-best results are underlined. The \textit{average rank} denotes the mean calculated across all seven datasets for the rank among all baselines pertaining to a specific metric.  OOM means out of memory.}
    \resizebox{2\columnwidth}{!}{
          \begin{tabular}{c|c|c|c|c|c|c|c|c|c}
            \toprule[0.8pt]
                                               & \textbf{Metric} & \textbf{Cora}       & \textbf{Citeseer}   & \textbf{Pubmed}     & \textbf{ACM}        & \textbf{Blog}       & \textbf{Flickr}     & \textbf{Reddit} & \cellcolor{lightgray}{\textit{ Average Rank}}     \\ 
            \midrule[0.8pt]
            \multirow{3}{*}{\textbf{DGI}}     & ARI             & 50.77±1.49          & 45.49±0.49          & 26.70±1.27          & 74.71±0.51    & 14.57±1.41          & 0.60±0.08           & OOM   &      6.0         \\   
                                              & NMI             & 56.63±1.19          & 44.63±0.45          & 28.96±1.68          & 69.67±0.57    & 24.32±1.47          & 1.31±0.19           & OOM     &      6.0       \\   
                                              & ACC             & 71.54±1.18          & 69.27±0.37          & 66.02±0.69          & 90.72±0.02    & 38.33±2.44          & 14.59±0.27          & OOM      &      6.0      \\ \hline
            \multirow{3}{*}{\textbf{MVGRL}}   & ARI             & {\underline{52.85±1.71}}    & 42.69±0.99          & 28.06±1.02          & 70.48±2.55          & 4.61±3.11           & 5.56±1.28           & OOM     &      5.1       \\   
                                              & NMI             & 58.10±0.94    & 43.01±0.91          & 31.14±1.14          & 65.84±2.08          & 9.26±8.10           & 14.38±3.32          & OOM      &     5.1       \\   
                                              & ACC             & 72.52±1.03          & 68.06±0.62          & 66.22±0.55          & 43.48±3.74          & 27.12±5.26          & 24.88±2.88    & OOM       &      6.9     \\ \hline
            \multirow{3}{*}{\textbf{SENet}}   & ARI             & 49.47±1.14          & 42.84±1.79          & 25.17±1.98          & 50.84±4.99          & NA                  & NA                  & OOM       &    9.4       \\   
                                              & NMI             & 55.65±0.32          & 42.11±1.48          & 26.70±2.72          & 47.65±4.69          & NA                  & NA                  & OOM      &      9.9      \\   
                                              & ACC             & 72.59±0.46          & 67.72±1.58          & 63.87±1.68          & 80.57±2.35          & NA                  & NA                  & OOM      &     9.3       \\ \hline
            \multirow{3}{*}{\textbf{AGE}}     & ARI             & 51.19±0.58         & 45.74±0.96          & 32.67±0.04          & {\underline{ 78.15±0.09}}           & 24.8±0.29           & 16.51±0.84           &        OOM     &  {\underline{ 3.4}}    \\   
                                              & NMI             & {\underline{ 58.18±0.21}}          & 45.06±0.43          & 31.01±0.15          & {\underline{ 72.6±0.11}}          & 33.46±0.14           & 27.88±0.76           &         OOM   &   3.6    \\   
                                              & ACC             & 74.13±0.22          & 70.06±0.72          & 70.78±0.05          & {\underline{ 92.13±0.03}}          & 54.98±0.14          & 39.16±01.79          &       OOM    &   {\underline{ 3.1 }}    \\ \hline
            \multirow{3}{*}{\textbf{AGC}}     & ARI             & 44.76±0.02          & 41.60±0.11          & 31.71±0.02          & 51.23±0.07          & 3.12±2.53           & 3.76±0.27           & OOM      &     8.1       \\   
                                              & NMI             & 53.49±0.02          & 41.00±0.09          & 31.51±0.02          & 49.58±0.08          & 9.82±2.75           & 9.55±0.32           & OOM      &     7.7       \\   
                                              & ACC             & 68.91±0.01          & 66.99±0.07          & 70.16±0.01          & 79.93±0.03          & 27.70±4.72          & 22.35±0.30          & OOM     &     7.6        \\ \hline
            \multirow{3}{*}{\textbf{DAEGC}}   & ARI             & 44.19±1.40          & 41.73±2.35          & OOM                 & 70.71±4.90          & 3.16±0.52           & 2.89±1.76           & OOM      &     8.6       \\   
                                              & NMI             & 49.10±0.85          & 41.40±1.57          & OOM                 & 65.91±3.59          & 4.45±1.01           & 5.44±3.32           & OOM    &      9.0        \\   
                                              & ACC             & 67.36±1.51          & 66.74±1.75          & OOM                 & 88.92±2.47          & 25.44±1.13          & 19.35±2.33          & OOM    &     8.9         \\ \hline
            \multirow{3}{*}{\textbf{SDCN}}    & ARI             & 26.85±6.78          & 24.10±2.60          & 18.50±3.75          & 56.41±9.48          & 4.04±4.16           & 2.10±1.43           & 16.12±3.01    &   9.7    \\   
                                              & NMI             & 33.25±4.06          & 26.26±2.29          & 20.75±2.72          & 53.08±6.94          & 9.21±5.44           & 4.88±1.62           & 39.50±2.40    &   10.1    \\   
                                              & ACC             & 52.41±5.64          & 52.30±2.20          & 59.80±3.95          & 80.64±8.31          & 26.43±5.29          & 17.98±2.35          & 28.04±3.05    & 9.9      \\ \hline
            \multirow{3}{*}{\textbf{AGCN}}    & ARI             & 24.26±4.73          & 25.64±8.81          & 23.78±2.47          & 45.88±11.58         & 9.20±4.18           & 0.68±1.34           & {\underline{ 52.79±3.31}} &  9.9  \\   
                                              & NMI             & 32.69±4.85          & 28.92±7.04          & 26.60±1.67          & 45.70±9.12          & 17.50±4.53          & 2.58±3.75           & 68.73±0.90       &  10.0  \\   
                                              & ACC             & 51.33±4.23          & 51.58±9.28          & 63.81±2.13          & 70.87±10.81         & 32.43±4.91          & 13.39±2.39          & {\underline{ 57.76±2.86}}  & 10.0  \\ \hline
            \multirow{3}{*}{\textbf{DFCN}}    & ARI             & 16.24±1.63          & 35.85±3.63          & 25.81±1.87          & 53.73±3.81          & 4.80±1.21           & 0.11±0.03           & OOM     &     10.0        \\   
                                              & NMI             & 26.67±1.51          & 36.52±3.24          & 28.06±2.10          & 51.18±2.81          & 12.50±1.45          & 0.72±0.25           & OOM    &       10.1       \\   
                                              & ACC             & 41.10±1.93          & 62.38±2.97          & 64.50±1.22          & 81.66±1.84          & 28.10±0.61          & 13.04±0.27          & OOM     &      9.7       \\ \hline
            \multirow{3}{*}{\textbf{DCRN}}    & ARI             & 24.22±6.91          & 44.76±0.43          & 33.02±0.31          & 69.72±0.33          & 2.68±0.26           & 1.40±0.61           & OOM      &      7.9      \\   
                                              & NMI             & 39.00±5.07          & 42.70±0.62          & \textbf{33.18±0.35} & 64.38±0.35          & 4.62±0.57           & 2.10±0.61           & OOM     &      7.6       \\   
                                              & ACC             & 54.49±5.33          & 68.63±0.43          & 70.21±0.27          & 88.83±0.14          & 24.47±0.20          & 15.36±0.85          & OOM     &      7.6       \\ \hline
            \multirow{3}{*}{\textbf{SEComm}}  & ARI             & 44.13±4.12          & 36.85±6.58          & 15.89±7.42          & 67.87±2.36          & 17.94±2.56          & 7.22±3.43     & OOM          &    7.7    \\   
                                              & NMI             & 50.15±2.44          & 37.43±5.13          & 18.74±6.01          & 63.21±1.85          & 25.55±3.10          & 13.62±3.90          & OOM        &     7.9      \\   
                                              & ACC             & 67.47±4.40          & 63.43±6.51          & 55.49±7.94          & 87.91±1.01          & 42.42±4.34          & 23.56±4.76          & OOM     &      7.6       \\ \hline
            \multirow{3}{*}{\textbf{AGC-DRR}} & ARI             & 3.70±0.90           & 1.62±0.25           & 0.37±0.56           & 7.39±2.83           & 0.00±0.00           & 0.00±0.00           & OOM     &     13.0        \\   
                                              & NMI             & 7.27±1.06           & 4.56±0.52           & 2.33±1.61           & 6.36±2.41           & 0.00±0.00           & 0.00±0.00           & OOM      &     13.0      \\   
                                              & ACC             & 30.03±1.77          & 25.07±0.81          & 41.88±1.87          & 47.24±2.42          & 17.73±0.00          & 11.72±0.00          & OOM     &     12.9        \\ \hline
            \multirow{3}{*}{\textbf{GCC}}     & ARI             & 51.2±0.00            & {\underline{ 46.63±0.10}}    & {\underline{ 33.24±0.00}}    & 47.70±0.00          & 21.20±0.00         & 3.52±0.24           & 43.57±4.39     &  4.9    \\   
                                              & NMI             & \textbf{58.9±0.00}   & \textbf{45.26±0.10} & {\underline{ 32.29±0.00}}    & 54.31±0.00          & 34.61±0.00         & 22.88±0.65    & {\underline{ 68.88±1.27}}  & {\underline{ 3.1}}  \\   
                                              & ACC             & {\underline{ 74.2±0.00}}      & {\underline{ 70.55±0.06}}    & {\underline{ 70.82±0.00}}    & 65.26±0.00          & 50.90±0.00         & 24.80±0.36          & 56.20±2.07      &  4.4   \\ \hline
            \multirow{3}{*}{\textbf{SUBLIME}} & ARI             & 46.59±2.48          & 42.41±0.16          & OOM                 & 64.05±1.56          & {\underline{ 45.96±1.78}}    & {\underline{ 28.34±0.59}}          & OOM      &     6.6       \\   
                                              & NMI             & 51.30±1.65          & 43.12±0.10          & OOM                 & 60.52±1.34          & {\underline{ 63.72±1.21}}     & {\underline{ 58.86±0.53}}           & OOM   &       6.3        \\   
                                              & ACC             & 67.25±2.55          & 67.06±0.16          & OOM                 & 85.75±0.76          & {\underline{ 75.98±0.89}}     & {\underline{ 64.22±0.78}}          & OOM    &       6.9       \\ 
            \midrule[0.8pt]
            \multirow{3}{*}{\textbf{\model}}    & ARI             & \textbf{53.91±0.23} & \textbf{48.19±0.25} & \textbf{33.75±0.04} & \textbf{80.10±0.49} & \textbf{81.07±0.11} & \textbf{60.59±0.12} & \textbf{66.23±1.05} & \textbf{1.0}\\   
                                              & NMI             & 57.63±0.14          & {\underline{ 45.24±0.24}}    & 31.75±0.04          & \textbf{74.57±0.55} & \textbf{78.43±0.12} & \textbf{62.49±0.13} & \textbf{77.51±0.31}& \textbf{1.9}\\   
                                              & ACC             & \textbf{74.33±0.19} & \textbf{72.38±0.18} & \textbf{71.34±0.04} & \textbf{92.90±0.18} & \textbf{91.49±0.05} & \textbf{75.40±0.17} & \textbf{70.95±1.00} & \textbf{1.0} \\ 
            \bottomrule[0.8pt]
            \end{tabular}
    }
    \label{tab:results}
\end{table*}

\begin{table}[t]
    \small
    \caption{Dataset Statistics. Hom. is short for homophily.}
    \resizebox{.48\textwidth}{!}{
        \begin{tabular}{c|r|r|r|r|r}
            \toprule[0.8pt]
            \textbf{Dataset}       & \textbf{Node}    & \multicolumn{1}{c|}{\textbf{Edge}}        & \textbf{Feat}   & \textbf{Cluster} & \textbf{Hom.} \\ \hline
            \textbf{Cora}           & 2,708   & 5,429       & 1,433  & 7    & 0.81 \\
            \textbf{Citeseer}       & 3,327   & 4,732       & 3,703  & 6    & 0.74 \\
            \textbf{ACM}            & 3,025   & 8,117       & 13,128 & 3    & 0.82 \\
            \textbf{Pubmed}         & 19,717  & 88,651      & 500    & 3    & 0.80 \\
            \textbf{BlogCatalog}  & 5,196    & 343,486     & 8,189  & 6    & 0.40 \\
            \textbf{Flickr}       & 7,575   & 479,476     & 12,047 & 9    & 0.24 \\
            \textbf{Reddit}       & 232,965 & 114,615,892 & 602    & 41   & 0.76 \\

            \bottomrule[0.8pt]
        \end{tabular}
    }
    \label{tab:datasets}
    
\end{table}

\subsection{Datasets}
In our evaluation, we employ a comprehensive set of seven publicly available datasets, comprising four citation networks and three social networks. For detailed information, please refer to Table \ref{tab:datasets}.

\textbf{Citation Networks}. We evaluate our work on four citation network datasets: Cora~\cite{pubmed}, Citeseer~\cite{pubmed}, ACM~\cite{acm}, and Pubmed~\cite{pubmed}. The nodes in Cora and Citeseer represent papers attributed with binary word vectors, while the paper nodes in Pubmed are represented with tf-idf weighted word vectors. The label of a paper indicates its subject and papers are connected if there is a citation relationship between them.


\textbf{Social Networks}. The evaluation is further conducted on three social network datasets: BlogCatalog (Blog)~\cite{blogflickr}, Flickr~\cite{blogflickr}, and Reddit~\cite{reddit}. The nodes in BlogCatalog and Flickr represent users of the respective websites where they share blogs and images, respectively. Relationships between users are indicated by connections among the nodes. The labels for nodes in Blog are topics of user blogs, while in Flickr, they are user interest groups.

\subsection{Baselines}
We compare the proposed \model method with four groups of baseline methods, whose details can be found in related works:

\textbf{Unsupervised GNNs with Traditional Clustering}. This group of methods first use GNNs to train node representations and then employs traditional clustering methods. Here we choose several widely used self-supervised node representation learning methods for comparison, i.e., DGI~\cite{velickovic2019deep}, MVGRL~\cite{hassani2020contrastive} and AGE~\cite{AGE} followed by K-means clustering. 
We also compare with two GNNs that learn clustering-friendly embedding, including SENet~\cite{zhang2021spectral} and AGC~\cite{zhang2019attributed}.

\textbf{Unsupervised Graph Structure Learning with Traditional Clustering}. As most existing GSL methods are semi-supervised, we compare with the representative unsupervised method SUBLIME~\cite{liu2022towards} which maximizes the agreement between the anchor graph and the learned graph in a contrastive mode. We compare it to show the advantages of considering clustering results for structure learning in graph clustering scenarios.

\textbf{Two-stage Graph Clustering Methods}. Most two-stage methods adopt the self-training schema proposed by DEC~\cite{DEC} to train the clustering  with pre-trained node representations. Among them, DAEGC~\cite{DAEGC}, AGCN~\cite{peng2021attention}, SDCN~\cite{bo2020structural}, DFCN~\cite{DFCN} and DCRN~\cite{DCRN} are the latest representative methods which we choose as baselines.

\textbf{One-stage Graph Clustering Methods}. Recently, the one-stage graph clustering methods have achieved superior performance. We also compare with the following one-stage graph clustering methods, namely  SEComm~\cite{SEComm}, GCC~\cite{fettal2022efficient}, and AGC-DRR~\cite{gong2022attributed}.

\subsection{Experimental Settings}

\textbf{Metric}. In the experiments, we evaluate the performance of graph clustering using the Accuracy (ACC), Normalized Mutual Information (NMI), and Adjusted Rand Index (ARI) as metrics, which have been widely used in prior studies~\cite{gan2020data,peng2021attention,bo2020structural,DFCN,DCRN}. The larger these metrics, the better the performance of graph clustering. To ensure the validity of the results, all experiments are run 3 times with different seeds. The mean and standard deviation are reported.

\textbf{Parameter settings}. We set the dimensionality of node representation to be 500 and use a learning rate of 0.001. The network is optimized with the Adam optimizer~\cite{adam}.
In terms of the hyperparameters for the clustering module, we vary the ratio $\gamma$ of high-confidence nodes extracted in each cluster from 0.1 to 1, the intra-cluster edge recovery ratio $\xi$ from 0.1 to 1, the inter-cluster edge removal ratio $\eta$ from 0.001 to 0.01, and max GSL epoch denoted as \textbf{Ep.} from 1 to 10. The best hyperparameters are selected for each dataset after exploring the range. 
The number of propagation times, $l$, is set to 8 for Cora, 3 for Citeseer and ACM, 35 for Pubmed, 1 for BlogCatalog and Flickr, and 3 for Reddit. $\kappa$ is uniformly set to 2/3 to achieve the best low-pass ability~\cite{AGE}.
For the baselines, we follow the parameter configuration given by the authors in their papers or GitHub repositories.
All experiments were run on a single Linux server with a single RTX3090 24G GPU.
Our source code is available at \url{https://github.com/galogm/HoLe}.

\subsection{Performance and Analysis (RQ1)}
Table \ref{tab:results} illustrates the experimental results of all baselines and our work. 
From the results, we have the following observations:

\db{Compared to other baselines without incorporating structure learning, the proposed \model method demonstrates superior performance on citation networks characterized by high initial homophily}. \model consistently ranks first or second in most metrics with the exception of NMI on a few datasets. The one-stage method GCC consistently ranks second across several datasets, while unsupervised GNNs combined with traditional clustering methods exhibit stable performance and rank in the third tier. Meanwhile, most two-stage methods, such as SDCN~\cite{bo2020structural}, show significant variance with unsatisfactory performance.

\db{In regards to the social network datasets with relatively low initial homophily, \model outperforms most baselines without structure learning by a substantial margin across all metrics.} 
In comparison, the performance of both two-stage methods and unsupervised GNNs with traditional clustering significantly drops compared to their performance on the citation networks. 
However, some one-stage methods like GCC and SEComm have achieved relatively better performance. 
For the baselines that do not incorporate structure learning and perform clustering on the suboptimal structures, they achieve relatively poor performance as expected.
In contrast, the \model method learns clustering along with structure learning and effectively enhance the homophily of the structure, achieves more significant improvements on social networks where initial structures are more likely to be suboptimal.

\db{Compared to SUBLIME, which is a representative unsupervised GSL method, our method also exhibits great advantages}. \model outperforms it by over 5\% on all citation networks and even more on social networks.
This highlights the limitations of not leveraging the clustering results into the structure learning.

\begin{table}[t]

    \small
    \caption{The efficacy of different modules within the GSL component. HCE is short for Hierarchical Correlation Estimation.}
    \label{tab:modgsl}
    \resizebox{.48\textwidth}{!}{
        \begin{tabular}{ll|ccc|ccc}
            \toprule[0.8pt]
            \multicolumn{2}{c|}{\multirow{2}{3em}{\textbf{Modules}}} & \multicolumn{3}{c|}{\textbf{Cora}} & \multicolumn{3}{c}{\textbf{Citeseer}}                                         \\

                                                                     &                                    & ARI                                   & NMI   & ACC   & ARI   & NMI   & ACC   \\\hline
            \textbf{w/o}                                             & \textbf{GSL}                       & 50.26                                 & 55.37 & 71.71 & 45.70 & 44.50 & 69.33 \\
            \textbf{w/o}                                             & \textbf{ERC}                       & 48.96                                 & 54.37 & 72.38 & 47.88 & 44.92 & 72.17 \\
            \textbf{w/o}                                             & \textbf{ERM }                      & 51.39                                 & 56.78 & 73.54 & 47.92 & 44.98 & 72.20 \\
            \textbf{w/o}                                             & \textbf{HCE}                    &                50.48                       &   56.65    &   73.59    &   47.44    &    44.57   &     71.79  \\\hline
            \textbf{w/}                                              & \textbf{GSL}                       & \textbf{53.91}                                 & \textbf{57.63} & \textbf{74.33} & \textbf{48.19} & \textbf{45.24} & \textbf{72.38} \\

            \bottomrule[0.8pt]
        \end{tabular}
    }
   
    \label{tab:ablation}
\end{table}

\subsection{Ablation Studies (RQ2)}
\label{sec:gnn}

In this section, we examine the contributions of different components in HoLe. Table~\ref{tab:modgsl} reports the results. We make three major observations. \db{Without the structure learning branch, the resultant variant (w/o GSL) performs poorly on 6 evaluation scenarios}. This observation highlights our motivation to integrate structure learning with graph clustering. \db{HoLe benefits from the cluster-oriented hierarchical correlation estimation (HCE)}. In two datasets, HoLe consistently outperforms the "w/o HCE" variant with an obvious margin, validating the effectiveness of our HCE proposal. \db{ Jointly considering the intra- and inter-cluster sparsification techniques, HoLe achieves the best results}. In Table~\ref{tab:modgsl}, HoLe performs consistently better than two variants (i.e., w/o ERC and w/o ERM) in all cases, indicating the reciprocal effects of utilizing both intra- and inter-cluster edge operations.

\begin{figure}[t]
    \centering
    
    \begin{subfigure}[]{0.32\textwidth}
         \includegraphics[width=\textwidth]{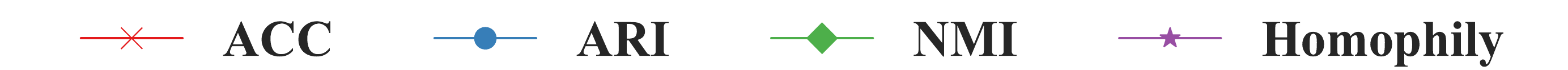}
         \label{fig:gcs-acc-homo-legend}
    \end{subfigure}
    \begin{subfigure}[]{0.18\textwidth}
         \includegraphics[width=\textwidth]{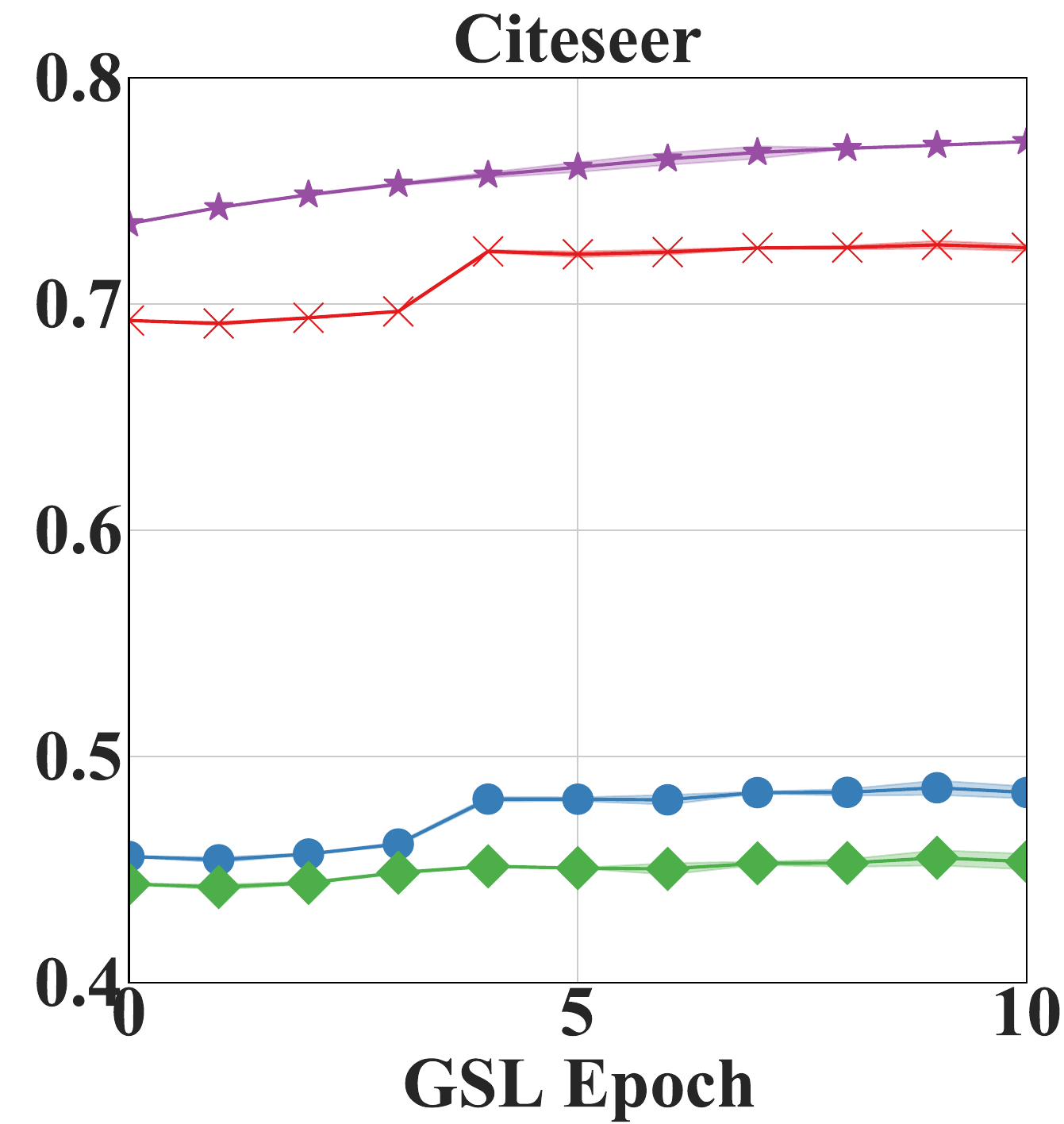}
         \label{fig:gcs-acc-homo-cite}
    \end{subfigure}
    \begin{subfigure}[]{0.18\textwidth}
         \includegraphics[width=\textwidth]{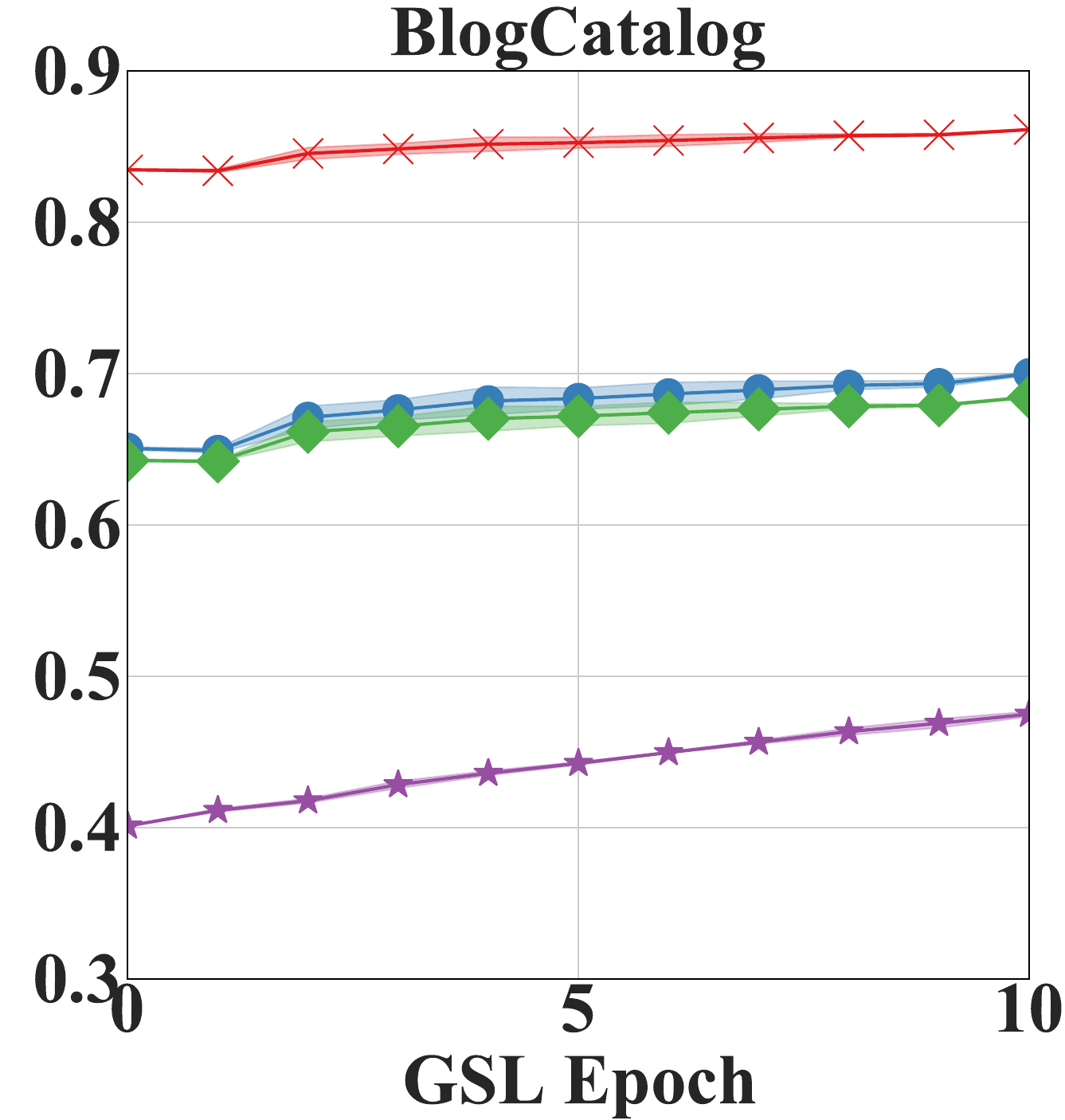}
         \label{fig:gcs-acc-homo-blog}
    \end{subfigure}
    \caption{Enhanced homophily leads to improved clustering performance on the Citeseer and BlogCatalog datasets.}
    \label{fig:enhancement-homophily}
\end{figure}

We also test the correlations between the homophily score of the learned structure and the graph clustering results. From Figure~\ref{fig:enhancement-homophily}, we observe that \db{The homophily learned by HoLe exhibit strong alignment with standard clustering metrics}. As the structure training epoch progresses, both the homophily values and the clustering scores (ACC, ARI, and NMI) consistently improve. These results further confirm the conclusion drawn from the sanity check conducted in Section~\ref{sec:data-analysis}, validating the enhancement of graph clustering through the improvement of homophily.

\begin{figure}[t]
    \centering
    \includegraphics[width=0.48\textwidth]{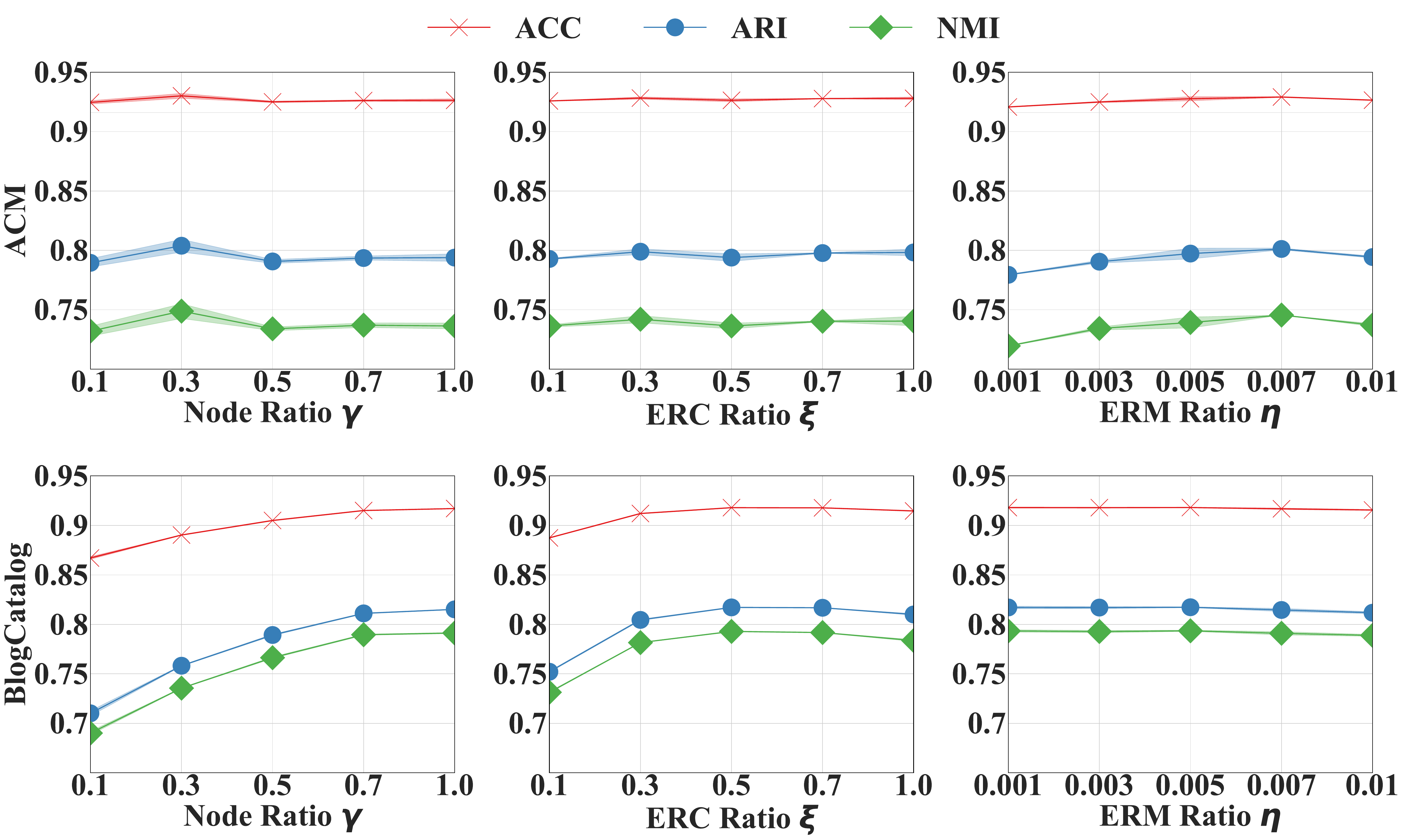}
    \caption{Hyperparameter analysis of inter-cluster edge removal (ERM) ratio $\eta$, intra-cluster edge recovery (ERC) ratio $\xi$, and high-confidence node extraction ratio $\gamma$. }
    \label{fig:param-analysis}
\end{figure}

\subsection{Sensitivity of Hyperparameters (RQ3)}
\label{sec:hyper-par}
In this section, experiments are conducted to explore the influence of the hyperparameters of the structure learning module.

As shown in Figure \ref{fig:param-analysis}, \db{the effects of individual hyperparameters on clustering performance demonstrate diverse change trends across different datasets with a relatively narrow fluctuation range.}
In citation networks characterized by high initial homophily, the ERC ratio $\xi$ generally exerts a minimal effect on clustering performance, while increasing the ERM ratio $\eta$ has a significant impact.
Conversely, social networks with low initial homophily demonstrate higher sensitivity to the $\xi$ ratio but are less affected by $\eta$.
This phenomenon can be attributed to the following explanation: In networks with high initial homophily, the addition of intra-cluster edges does not substantially improve performance since the neighborhoods already have numerous similar nodes.
In contrast, networks with low initial homophily benefit significantly from the addition of intra-cluster edges, enhancing homophily, while removing inter-cluster edges does not contribute to nodes having more neighbors within the same cluster.

\begin{figure}[t]
    \centering    
    \begin{subfigure}[]{0.115\textwidth}
         \includegraphics[width=\textwidth]{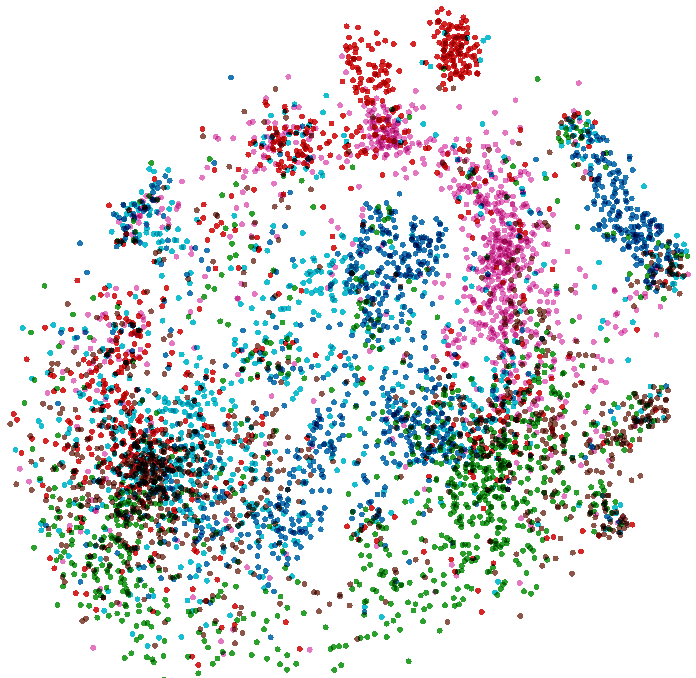}
         \caption{MVGRL}
         \label{fig:tsne_MVGRL_blog}
    \end{subfigure}
    \begin{subfigure}[]{0.115\textwidth}
         \includegraphics[width=\textwidth]{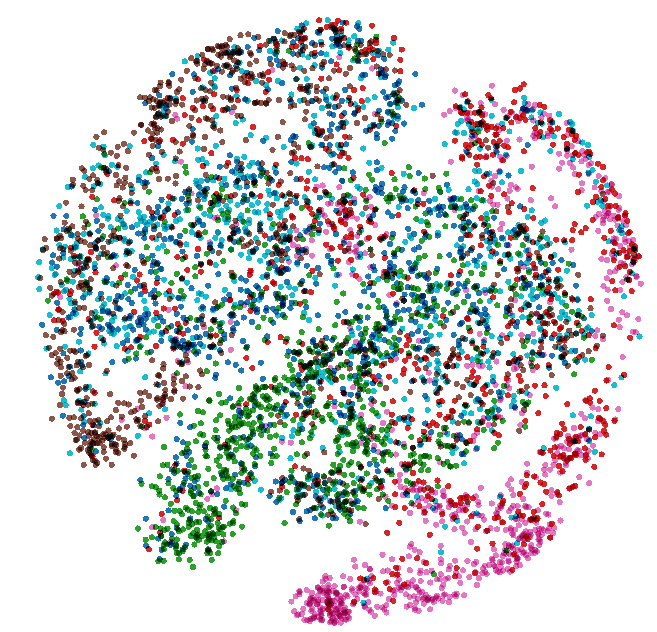}
         \caption{AGCN}
         \label{fig:tsne_AGCN_blog}
    \end{subfigure}
    \begin{subfigure}[]{0.115\textwidth}
         \includegraphics[width=\textwidth]{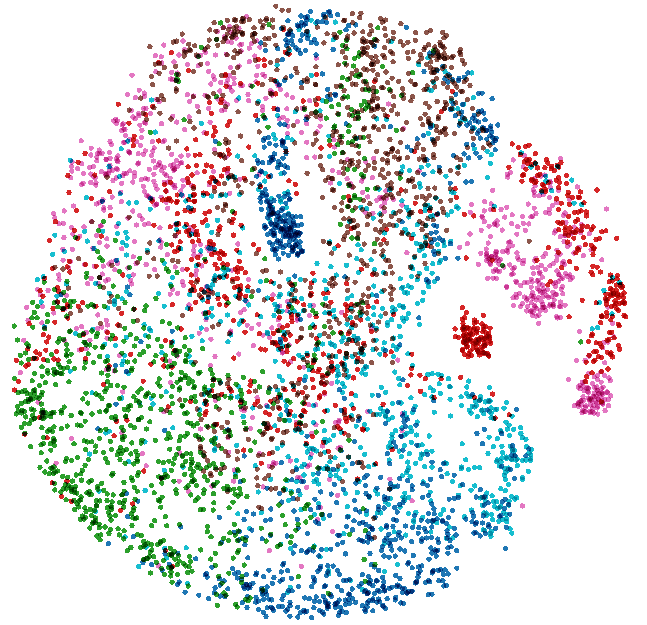}
         \caption{SEComm}
         \label{fig:tsne_SEComm_blog}
    \end{subfigure}\\
    \begin{subfigure}[]{0.115\textwidth}
         \includegraphics[width=\textwidth]{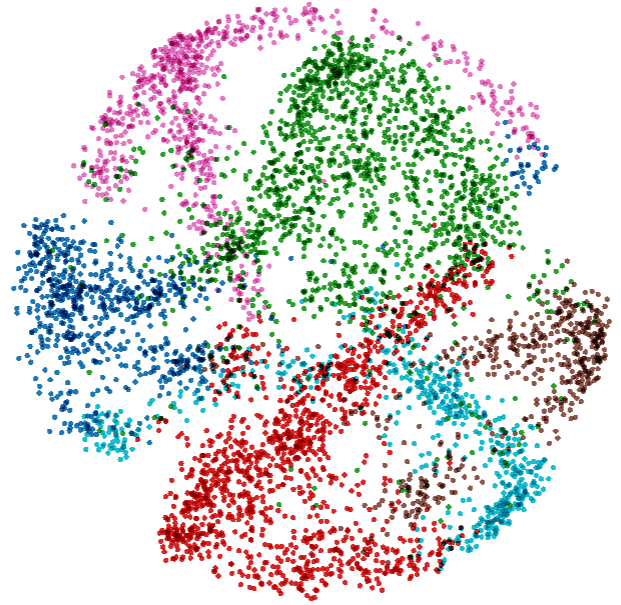}
         \caption{GCC}
         \label{fig:tsne_gcc_blog}
    \end{subfigure}
    \begin{subfigure}[]{0.115\textwidth}
         \includegraphics[width=\textwidth]{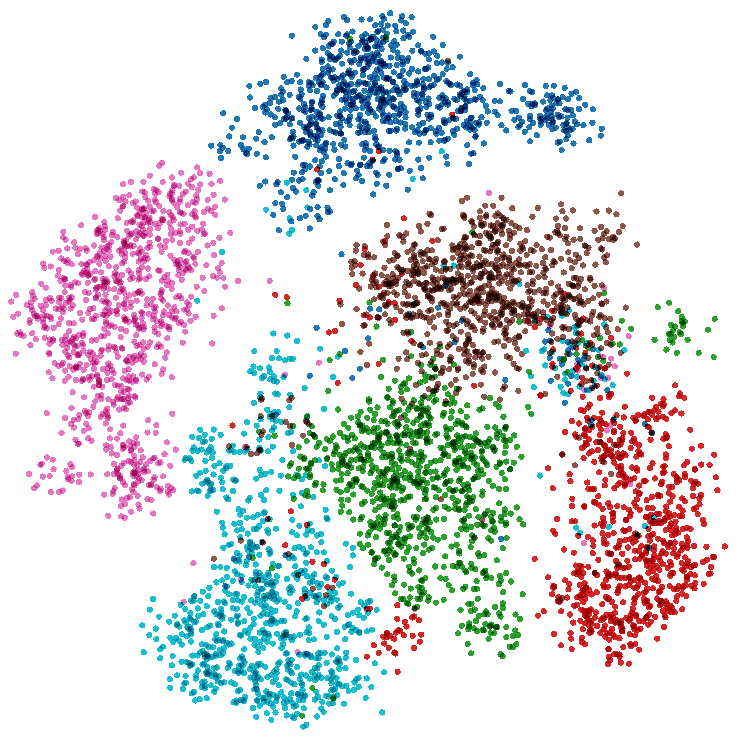}
         \caption{SUBLIME}
         \label{fig:tsne_SUBLIME_blog}
    \end{subfigure}
    \begin{subfigure}[]{0.115\textwidth}
         \includegraphics[width=\textwidth]{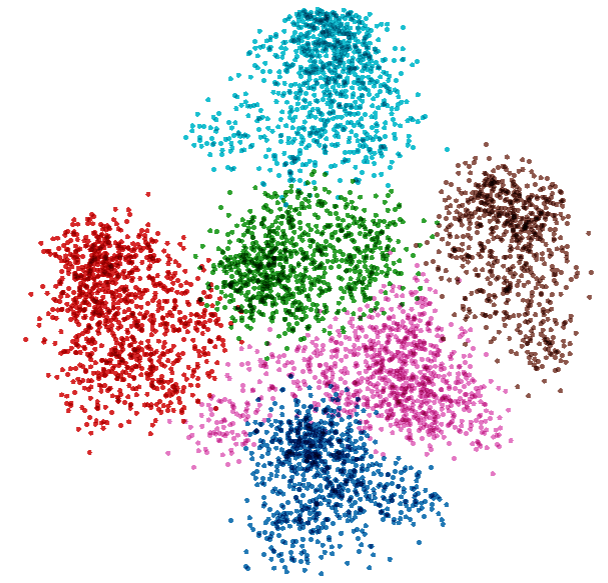}
         \caption{\model~(Ours)}
         \label{fig:tsne_gcs_blog}
    \end{subfigure}
    \caption{Visualization of different models on BlogCatalog by $t$-SNE.}
    \label{fig:visual}
    
\end{figure}

\subsection{Visualization}

To visually illustrate the benefits of our model's clustering effect, we employ the $t$-SNE algorithm~\cite{van2014accelerating} to visualize the clustering outcomes. 
As depicted in Figure \ref{fig:visual},
\db{\model surpasses all the selected baselines by distinguishing the clusters more effectively, as indicated by minimized overlapping areas and a higher level of aggregation among nodes in the same group.}

\section{Conclusion}
In this paper, we introduce \titlename~(\model). Our proposal is motivated by the empirical observation that subtle enhancements to the degree of homophily within the graph structure can lead to significant improvements in GNN-based graph clustering.
To achieve this, our approach incorporates hierarchical correlation estimation and cluster-aware sparsification modules to accurately estimate the relationships between pairs of nodes.
These modules are guided by both the latent space and the clustering space, allowing us to generate a sparsified structure that enhances homophily.
Through joint optimization, we demonstrate the superiority of \model over state-of-the-art baselines on 7 benchmark datasets.

In our future research, we are going to extend the homophily-enhanced structure learning module to large-scale graphs. Our focus will be on addressing the scalability challenges faced by graph clustering, ensuring practical applicability in real-world scenarios.

\begin{acks}
  This research was supported by  National Natural Science Foundation of China (Grant No: 62106221, 61972349), Zhejiang Provincial Natural Science Foundation of China under Grant No. LTGG23F030005, Ningbo Natural Science Foundation (Grant No: 2022J183).
\end{acks}

\bibliographystyle{ACM-Reference-Format}
\balance
\bibliography{references}


\end{document}